\RequirePackage{fix-cm}
\documentclass[smallextended]{svjour3}       
\smartqed  
\usepackage{url}
\usepackage{graphicx}
\usepackage[utf8]{inputenc}
\usepackage{multirow}
\usepackage{todonotes}
\usepackage[authoryear]{natbib}

\begin{document}

\title{A User-Study on Online Adaptation of Neural Machine Translation to Human Post-Edits
}

\titlerunning{A User-Study on Online Adaptation in NMT}        

\author{Sariya Karimova \and Patrick Simianer \and Stefan Riezler}

\authorrunning{Karimova, Simianer, Riezler} 

\institute{Sariya Karimova \and Patrick Simianer \and Stefan Riezler \at
              Heidelberg University - Department of Computational Linguistics \\
              69120 Heidelberg, Germany\\
              Tel.: +49622154-3245\\
              \email{\{karimova, simianer, riezler\}@cl.uni-heidelberg.de}      
    \and
    Sariya Karimova \at Kazan Federal University \\
    420008 Kazan, Russia
}

\date{Received: date / Accepted: date}

\maketitle

\begin{abstract}
The advantages of neural machine translation (NMT) have been extensively validated for offline translation of several language pairs for different domains of spoken and written language. However, research on interactive learning of NMT by adaptation to human post-edits has so far been confined to simulation experiments. We present the first user study on online adaptation of NMT to user post-edits in the domain of patent translation. Our study involves 29 human subjects (translation students) whose post-editing effort and translation quality were measured on about 4,500 interactions of a human post-editor and a machine translation system integrating an online adaptive learning algorithm. Our experimental results show a significant reduction of human post-editing effort due to online adaptation in NMT according to several evaluation metrics, including hTER, hBLEU, and KSMR. Furthermore, we found significant improvements in BLEU/TER between NMT outputs and professional translations in granted patents, providing further evidence for the advantages of online adaptive NMT in an interactive setup.
\end{abstract}


%
%

\section{Introduction}
\label{intro}
The attention-based encoder-decoder framework for neural machine translation (NMT) \citep{BahdanauETAL:15} has been shown to be advantageous over the well-established paradigm of phrase-based machine translation immediately after its inception. For example, significant improvements according to automatic and manual evaluation metrics could been shown in benchmark translation competitions for spoken language \citep{LuongManning:15} and written language \citep{JeanWMT:15,SennrichWMT:16}. These results have been investigated in-depth in analyses of the advantages of NMT along linguistic dimensions \citep{BentivogliETAL:16} and along different domains \citep{castilho2017neural}. In contrast, research on uses of NMT in online adaptation scenarios has so far been confined to simulations where the interactions of an NMT system with a human post-editor are simulated by a given set of static references \citep[][\emph{inter alia}]{WuebkerETAL:16,KnowlesKoehn:16,turchi2017continuous,peris2017interactive} or by a set of offline generated post-edits \citep{turchi2017continuous}. User-studies on the benefits of machine learning for adaptation of translation systems to human post-edits are rare, and to the best of our knowledge, such studies have so far been restricted to phrase-based machine translation \citep{DenkowskiHaCAT:14,GreenETAL:14,bentivogli2016evaluation,simianer2016post}.

We present a user study that analyzes 4,500 per-sentence interactions between an NMT system and 29 human post-editors. Our target domain are patents that have to be translated from English into German. Our goal is to quantify the mutual benefits of a system that immediately learns from user corrections, on the one hand by reducing human post-editing effort, and on the other hand by improving translation quality of the NMT output. In comparing post-editing of NMT outputs that are generated from systems with and without online adaptation, we find a significant reduction in post-editing effort for the former scenario according to the metrics of hTER (and hBLEU) between NMT outputs and human post-edits. This confirms findings that have been reported for user studies on online adaptation of phrase-based systems \citep{bentivogli2016evaluation, simianer2016post}. Moreover, we find significant improvements of post-editing effort for the online adaptation scenario with respect to metrics such as keyboard strokes and mouse clicks that have been used in computer-assisted translation.

We also attempt to quantify improvements in translation quality by measuring improvements in sentence-level BLEU+1 \citep{nakov2012optimizing} and TER between the iteratively refined NMT outputs and static human reference translations as found in granted patents. We find significant improvements with respect to both metrics, showing a domain adaptation effect due to online adaptation. This provides further evidence for the advantages of an online adaptive NMT system in an interactive setup.

The remainder of this paper is organized as follows: In Section \ref{related}, we discuss the related work. We briefly introduce the learning protocol of online adaptation, describe the tools and data, and the experimental design of our user study (Section \ref{experiments}). Experimental results will be discussed in Section \ref{results}. We conclude the paper by conclusions to be drawn from our experiments (Section \ref{conclusion}).

\section{Related Work}
\label{related}
The advantages of NMT and its challenges have been investigated from different angles in recent work \citep[][\emph{inter alia}]{W17-3204,E17-1100,E17-2045,macketanz2017machine,castilho2017neural, klubivcka2017fine,bentivogli2018neural,D17-1263,popovic2017comparing,forcada2017making,castilhocomparative,JunczysDowmunt2016IsNM,klubivcka2018quantitative,shterionovalphaempirical,burchardt2017linguistic}, 
however, studies on interactive NMT, especially user studies involving human post-edits of NMT outputs, have so far not been presented.

Online adaptation has been thoroughly researched since at least a decade, either by adding online discriminative learning techniques \citep[][\emph{inter alia}]{CesaBianchiETAL:08,MartinezGomezETAL:12,Lopez-SalcedoETAL:12,DenkowskiETAL:14,BertoldiETAL:14} to phrase-based MT systems, or adaptations to the generative components of the phrase-based framework \citep[][\emph{inter alia}]{NepveuETAL:04,OrtizMartinezETAL:10,HardtElming:10}. Recent studies applied the online adaptation framework to NMT, however, by simulating the interactive scenario by online learning from offline created human references or post-edits \citep{WuebkerETAL:16,KnowlesKoehn:16,turchi2017continuous,peris2017interactive}.

User-studies involving the generation of post-edits in an online interaction between translation system and post-editor have been presented as well, however, to the best of our knowledge these studies have been confined so far to phrase-based MT \citep{green2013efficacy,GreenETAL:14,DenkowskiHaCAT:14,bentivogli2016evaluation,simianer2016post}.\footnote{In addition, \cite{GreenETAL:14} -- one of the first user studies on online adaptation to post-edits -- performed system updates offline instead of online.}
The closest approach to our work is the study presented by \cite{bentivogli2016evaluation}. Similar to our work, they involve human subjects in an interactive post-editing scenario where the system learns online from user corrections. However, their study is confined to phrase-based machine translation, and differs from our study in choosing a within-subjects experimental design where the same translator post-edits the same document under either test condition (static/adaptive NMT). We use a more standard between-subjects design where each session is composed of different documents of comparable difficulty, each of which is translated under two different conditions (with and without online NMT adaptation), and post-edited by two different translators.

\section{Experimental Setup}
\label{experiments}
\subsection{Online Adaptation}

Online adaptation for NMT follows the online learning protocol shown in Figure \ref{fig:online-learning} that we adopted from \cite{BertoldiETAL:14}. The protocol assumes a global model $M_g$ that was trained on a large dataset of parallel data. This dataset does not necessarily come from the same domain as the data used in online adaptation. Online adaptation proceeds by performing online fine-tuning on a further set of patent documents $d$, resulting in a combined model $M_{g+d}$. This is done by invoking a sequence of $|d|$ interactions, where on each step a translation output $\hat y_t$ for an input source segment $x_t$ is produced by the NMT system, a post-edit $y_t$ is produced by the user, and a system update is performed by using the pair $(x_t,y_t)$ as supervision signal in online learning.

\begin{figure}
  \begin{description}
\item[\hspace{-0.5ex}] Train global model $M_g$
\item[\hspace{1ex}\textbf{for each}] document $d$ of $|d|$ segments
\begin{description}
\item[\hspace{1ex}\textbf{for each}] example $t=1,\ldots,|d|$
\begin{description}
\item[\textbf{1.}] Receive input sentence $x_t$
\item[\textbf{2.}] Output translation $\hat y_t$ 
\item[\textbf{3.}] Receive user post-edit $y_t$
\item[\textbf{4.}] Refine $M_{g+d}$ on pair $(x_t,y_t)$
\end{description}
\end{description}
\end{description}
	\caption{Online learning protocol for post-editing workflow}
	\label{fig:online-learning} 
\end{figure}

\subsection{NMT System}

The NMT system used in our experiments is based on the Lamtram toolkit \citep{neubig15lamtram}, which is built on the dynamic neural network library DyNet \citep{neubig2017dynet}. It implements an encoder-decoder architecture with attention mechanism. The settings in our experiments use dot product as attention type, together with attention feeding \citep{luong2015effective} where the context vector of the previous state is used as input to the decoder neural network. We trained recurrent neural networks (RNNs) with 2 layers consisting of 256 units, and a word representation layer of 128 units, on GPU. The chosen RNN architecture is a long short-term memory network \citep{hochreiter1997long}. As stochastic optimization method we used ADAM \citep{kingma2014adam} with a learning rate initialization to $0.001$. To prevent overfitting, we set the dropout rate to $0.5$ \citep{srivastava2014dropout}, and used a development set of 2k sentences for early stopping. Evaluation was performed after every 50k sentences. 

In order to use Lamtram as an interactive online learning platform, we needed to modify the tool to allow training and translation to take turns without having to reload the model parameters. 

For online adaptation we used stochastic gradient descent, with a learning
rate of $0.05$ and a dropout of $0.25$. For inference, the beam size was set to 10, and we tuned a word penalty parameter to adjust the lengths of the outputs, as well as a penalty for unknown words. These parameters were set to $0.85$ and $0.25$ respectively.

\subsection{User Interface}

\begin{figure}[t]
  \includegraphics[width=1\textwidth]{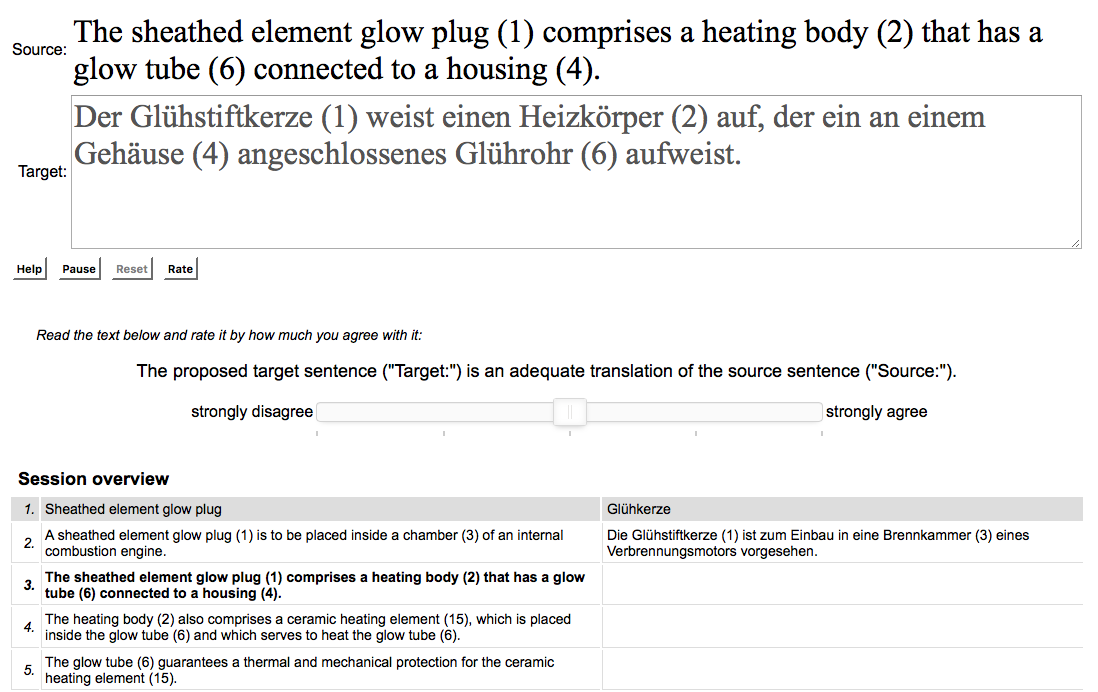}
\caption{User interface for post-editing}
\label{fig:interface}       
\end{figure}

Furthermore, we implemented a user interface that sends inputs via the network to a web client that renders the source and the proposed NMT for user post-editing, and records post-edits for learning. A screenshot of the interface is shown in Figure \ref{fig:interface}. From top to bottom, it shows the source, the post-editing field, and the slider used to collect human quality ratings of the NMT outputs before post-editing. Segments are complete patent abstracts in their
original order preluded by the respective patent's title.

\subsection{Data}

As training data we used $\sim$ 2M parallel sentences extracted from Europarl and News Commentary. Furthermore, offline fine-tuning was performed on $\sim$ 350k parallel sentences of in-domain data from PatTR\footnote{\url{http://www.cl.uni-heidelberg.de/statnlpgroup/pattr/}}. The translation direction is from English into German. 

Since patent claims and descriptions tend to be extremely complex and long,
they are not suitable for translation by non-experts. We therefore used titles and abstracts for both training and test. Development and test data are limited to documents, each consisting of a patent title and abstract, with an overall maximum length of 45 tokens per sentence. The data split was done by year and by family id to avoid any possible overlaps. The test data were automatically grouped into clusters by cosine similarity of their bag-of-words tf-idf source representations and length, to obtain clusters of related documents with an approximate source token count of 500, which is appropriate in a post-editing setup given the available time limit of 90 minutes. This way, each cluster contained the titles and complete abstracts of 3-5 documents. All data were preprocessed by tokenization, truecasing, and byte pair encoding \citep{sennrich2015neural} with a vocabulary size of 10k for source and target, respectively.

\subsection{Experimental Design}

Our post-editors were 29 master-level students at the Institute for Translation and Interpreting at Heidelberg University. The experiments were conducted during 8 post-editing sessions with a duration of 60-90 minutes each over the course of 5 days. All sessions took place in the same computer pool with the same hardware on each computer. 

In a real-world post-editing scenario translator usually has access to online available dictionaries, translation memories or other resources. However, in our specific case the patents which we used in our experiments are online available\footnote{\url{http://www.cl.uni-heidelberg.de/statnlpgroup/pattr/}}, indexed, and can be retrieved by concordance search\footnote{for example, on \url{https://www.linguee.com}}. The data for PatTR was extracted from the MAREC patent collection, which also can be found online\footnote{\url{http://www.ifs.tuwien.ac.at/imp/marec.shtml}}. Thus in order to avoid receiving copies of the original data instead of human post-edits, we had to compromise the experiment setup by restricting post-editor to use only online dictionaries, that do not crawl for parallel data\footnote{\url{http://www.dict.cc, http://dict.leo.org}}, and Wikipedia\footnote{\url{https://de.wikipedia.org}}.

In our experiments, we processed overall 105 documents, resulting in 2,209 source sentences.  The same documents were used to test the effect of the adaptation condition on human post-editing, however, in difference to \cite{bentivogli2016evaluation}, no document was seen by the same translator twice. We avoided on purpose a possible scenario assigning one document to the same student two times, once translated by a baseline and once by an adapted NMT system. This is to avoid a learning/memorizing effect for the second time a post-editor views and edits the same document. Instead, all documents were translated under two different translation conditions (with and without online adaptation) and post-edited by two different students.

To sum up the experimental setup, there were two automatic translations of the same 2,209 sentences (adaptive NMT + static NMT) and one post-edit for each condition.\footnote{A single document was used as an exam in the very last session and translated by all post-editors and without adaptation.} In total we collected 4,563 per-sentence measures (NMT adaptation false - 2,354 and true - 2,209) of (h)BLEU \citep{papineni2002bleu}, (h)TER \citep{snover2006study}, translation quality rating \citep{GrahamETAL:16}, keyboard strokes and mouse clicks \citep{barrachina2009statistical}, and post-editing time. 

\begin{table}[t]
\caption{Excerpt for the model coefficients for the used fixed effect of online NMT adaptation to the individual intercepts for response variable hBLEU; in the example, the global intercept has a value of 47.19 and the global slope lies at 6.73}
\label{tab:slopes}     
\begin{center} 
\begin{tabular}{lll}
\hline\noalign{\smallskip}
Random effect & Individual intercept & Individual slope \\
\noalign{\smallskip}\hline\noalign{\smallskip}
sentenceID\_15 & 39.64 & 10.67 \\
sentenceID\_16 & 49.86 & 19.52 \\
sentenceID\_17 & 53.12 & 23.73 \\
sentenceID\_18 & 53.39 & 11.75 \\
sentenceID\_19 & 66.55 & 20.39 \\
\noalign{\smallskip}\hline\noalign{\smallskip}
user\_A & 45.96 & 3.06 \\
user\_B & 53.66 & 10.63 \\
user\_C & 51.23 & 15.01 \\
user\_D & 37.77 & 5.06 \\
user\_E & 53.07 & 7.28 \\
user\_F & 54.10 & 8.97 \\
\noalign{\smallskip}\hline
\end{tabular}
\end{center}
\end{table}

\section{Analysis and Results}
\label{results}
\subsection{Statistical Analysis}

To analyze the results, we used linear mixed-effects models (LMEMs), implemented in the lme4 package \citep{bates2015fitting} in R \citep{2014r}. \cite{baayen2008mixed} introduced the usage of LMEMs for the analysis of repeated measurement data, enabling to resolve non-independencies by introducing sources of variation, by-subject and by-item variation, as random effects into the model. The general form of an LMEM can be described as the unconditional distribution of a vector of random effects $b$, and the conditional distribution of a vector-valued random response variable $Y$ given $b$, which are both multivariate Gaussian distributions \citep{bates2015fitting}. In matrix form, the LMEM can be expressed by the following formula: 
\begin{equation}
Y = X\beta + Zb + \epsilon,  
\end{equation}
where $\beta$ and $b$ are fixed-effects and random-effects vectors, $X$ and $Z$ are fixed-effects and random-effects design matrices, and $\epsilon$ is a vector of random errors.

In our application, the main fixed effect is the NMT adaptation condition (online adaptation to post-edits versus offline learning of a global model only), for which several response variables were measured. The random effects have differing intercepts as well as differing slopes. The granularity of the model is at the sentence level.
The observed response value for the $i$-th subject and $j$-th sentence, $y_{ij} \in Y$ (for example, a time measurement, hTER, hBLEU, etc.), is defined in our application of LMEMs as follows:
\begin{equation}
y_{ij} = \beta_0 + (\beta_1 + \beta_2)x_{ij} \\
+ b_{0i} + b_{0j} + (b_{1i} + b_{1j})z_{ij} + \epsilon_{ij}.  
\end{equation} 
The LMEM yields estimates of a global intercept $\beta_0$ (i.e., the expected mean value of a response variable when all slopes are equal to 0) and global slopes for the used fixed effects of NMT adaptation $\beta_1$ and dayID $\beta_2$. Thus in our experiments, a global intercept $\beta_0$ is an estimate for an average value for the measurements across all students and all sentences on the first day in the scenario without NMT adaptation. A global slope of the main fixed effect $\beta_1$ provides an estimate for the difference due to online NMT adaptation. The global slope $\beta_2$ of the dayID fixed effect estimates differences in response variables in consecutive sessions. Furthermore, we get random effect intercepts for subject $b_{0i}$ and sentence levels $b_{0j}$ (i.e., for each level we get that level's intercept's deviation from the global intercept) and random effect slopes within each user $b_{1i}$ and sentence level $b_{1j}$ (i.e., the degree to which a fixed effect  deviates from the global slope within a given level). Thus, each student and each sentence get its own individual intercept, or average value, in the scenario without NMT adaptation, as well as its own estimate for improvement due to online NMT adaptation. In our case, $(x_{ij}) = X$ being equal to $(z_{ij}) = Z$, is a design matrix of categorical variables with regard to the measurement for the $j$-th sentence and $i$-th subject and the respective predictor; $\epsilon_{ij}$ is an error term. 

\begin{table}[t]
\caption{Slopes for the used fixed effects of online NMT adaptation and dayID (index of day when consecutive sessions took place) to the global intercept; the response variable is post-editing time in ms, with a global intercept of 811.76 $\pm$ 43.47}
\label{tab:lmem_time}   
\begin{center}   
\begin{tabular}{ll}
\hline\noalign{\smallskip}
Fixed effect & Slope \\
\noalign{\smallskip}\hline\noalign{\smallskip}
NMT adaptation & -29.72 $\pm$ 27.90 \\
\noalign{\smallskip}\hline\noalign{\smallskip}
dayID\_2 & -135.81 $\pm$ 38.33 \\
dayID\_3 & -171.21 $\pm$ 38.30 \\
dayID\_4 & -221.24 $\pm$ 38.25 \\
dayID\_5 & -235.82 $\pm$ 35.30 \\ 
\noalign{\smallskip}\hline
\end{tabular}
\end{center}
\end{table}

We applied the idea of maximum random effects in our model by using random slopes for each random effect to account for different reactions of subjects and for different effects for items with regard to experimental conditions \citep{BarrETAL:13}. Table \ref{tab:slopes} illustrates the usefulness of the used maximum random effects structure by a sample of individual intercepts and respective slopes for each level of random effects in the LMEM for hBLEU. First, interpreting individual intercepts as average hBLEU values without NMT adaptation, we observe that both individual intercepts for sentences and post-editors differ from the global intercept, which is the average value of 47.19 found in measurements in the offline learning scenario. The high variance within individual intercepts of each random effect indicates varying difficulty of sentences to be translated and diverse preferences and experience of post-editors. Second, the individual slope values, interpreted as hBLEU improvements due to online NMT adaptation, make evident how much influence the effect of NMT adaptation has on each sentence and for each post-editor. In comparison to a global slope of 6.73, individually estimated random effect slopes vary significantly. This shows that different sentences are harder to improve and different post-editors react differently to manipulations to the NMT system.

Also, we found it useful to add the dayID as an additional fixed effect in the LMEM. The variable dayID labels the days in chronological order when consecutive sessions took place. It indicates the progress of post-editors through time and their improving experience with the NMT system and post-editing practice. An example is given in Table \ref{tab:lmem_time} which compares the speedup of the post-editing time in consecutive sessions to the improvement in post-editing time due to NMT adaptation. We can clearly see that the learning effect over time (shown in the reduction in post-editing time in consecutive sessions) has a larger impact than the effect of online NMT adaptation.  

\begin{table}[t]
\caption{Slope for the fixed effect of online NMT adaptation to the global intercept. LMEMs were built for hBLEU, TER/BLEU between NMT output (mt) and reference (ref), TER/BLEU between post-edit (pe) and reference, hTER, rating, count of keystrokes and mouse clicks, KSMR and time as response variables; significance of results was tested with likelihood ratio tests of the full model against the model without the independent variable of interest}
\label{tab:lmem} 
\begin{center}     
\begin{tabular}{llll}
\hline\noalign{\smallskip}
Response variable & Intercept & Slope & Significance \\
\noalign{\smallskip}\hline\noalign{\smallskip}
hBLEU (\%) & 47.19 $\pm$ 1.20 & +6.73 $\pm$ 1.01  & p \textless 0.001 \\
BLEU mt \& ref (\%) & 20.79 $\pm$ 0.97 & +1.76 $\pm$ 0.27 & p \textless 0.001 \\ 
BLEU pe \& ref (\%) & 22.76 $\pm$ 1.22 & +1.09 $\pm$ 0.63 & p \textless 0.1 \\
\noalign{\smallskip}\hline\noalign{\smallskip}
hTER (\%) & 35.45 $\pm$ 0.82 & -4.98 $\pm$ 0.72 & p \textless 0.001 \\
TER mt \& ref (\%) & 56.28 $\pm$ 0.97 &-0.95 $\pm$ 0.29 & p \textless 0.02 \\ 
TER pe \& ref (\%) & 54.86 $\pm$ 1.18 & -0.53 $\pm$ 0.55 & - \\
\noalign{\smallskip}\hline\noalign{\smallskip}
rating (0-100) & 44.23 $\pm$ 2.39 & +6.11 $\pm$ 1.42 & p \textless 0.001 \\
\noalign{\smallskip}\hline\noalign{\smallskip}
kbd+click (count) & 73.15 $\pm$ 4.76 & -12.13 $\pm$ 2.08 & p \textless 0.001 \\
KSMR (ratio) & 0.52 $\pm$ 0.02 & -0.07 $\pm$ 0.02 & p \textless 0.001 \\
\noalign{\smallskip}\hline\noalign{\smallskip}
time (ms) & 811.76 $\pm$ 43.47 & -29.72 $\pm$ 27.90 & - \\
\noalign{\smallskip}\hline
\end{tabular}
\end{center}
\end{table}

\subsection{Experimental Results}

Table \ref{tab:lmem} gives the central results of our analysis: We find significant improvements in post-editing effort due to online adaptation, shown in reduced hTER by nearly 5 points and improved hBLEU up to 6.73 points. Furthermore, online adaptation has a domain adaptation effect which leads to translation outputs which are closer to the static reference translations in the granted patents. This is shown in an increase of BLEU between NMT output and reference translation as well as BLEU between post-edit and reference. This agrees with the quality assessment (rating) that users had to give using a 100-point slider before they can start the post-editing process: Post-editors assess the quality of the NMT output at 6 points higher in case of online adaptation.  

\begin{figure}[t]
  \includegraphics[width=0.85\textwidth, page=2]{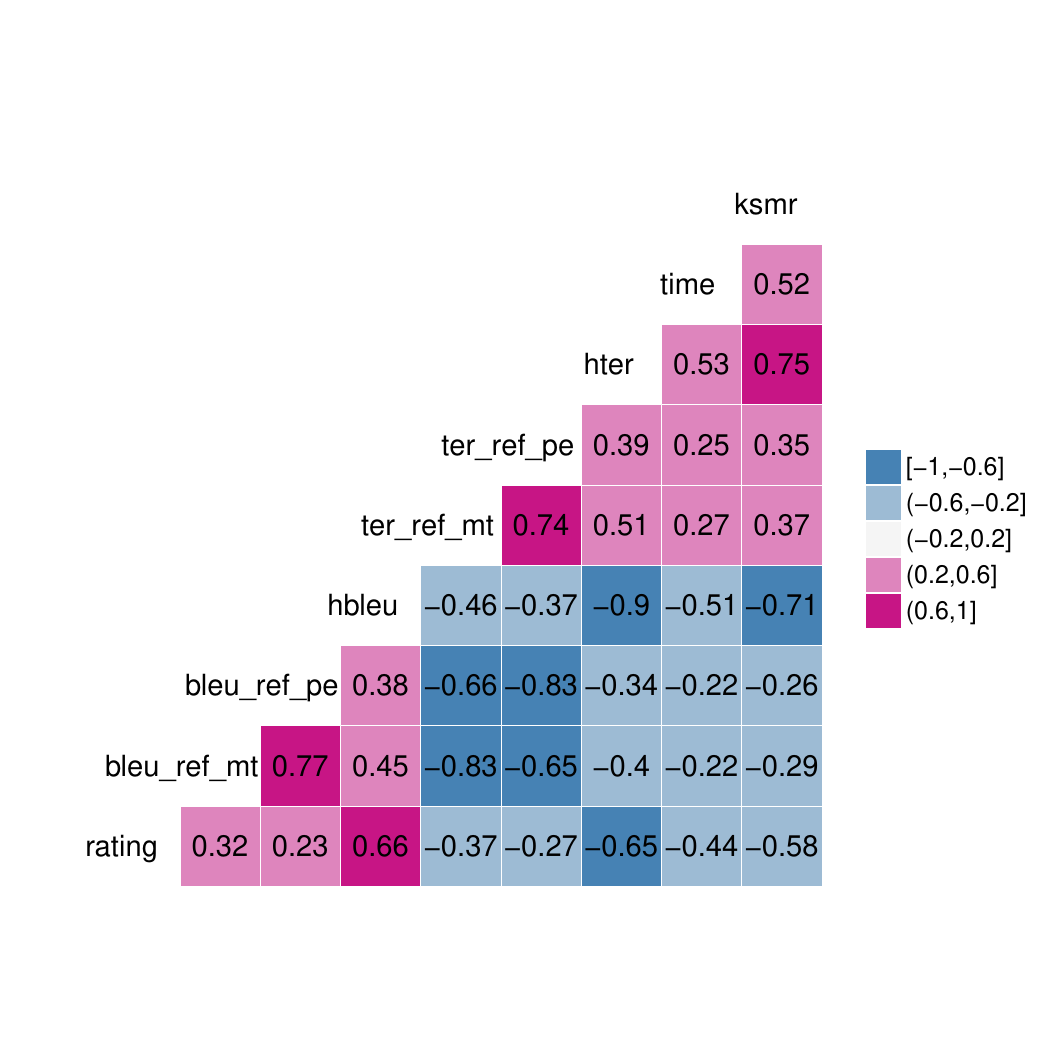}
\caption{Correlation matrix for pairwise Pearson correlation coefficients for KSMR, time, hTER, hBLEU, TER/BLEU between reference and post-edit, TER/BLEU between reference and NMT output, rating; strong positive correlations are marked with dark red; strong negative correlations by dark blue}
\label{fig:corr_matrix_P}       
\end{figure}

In measuring post-editing time, we normalized wall-clock time by the number of characters in a post-edit. The improvement in time is 30 ms less per character, corresponding to a nominal improvement of 3.7\%. We conjecture that the reason why we could not establish a significant improvement in post-editing time could be due to improved post-editors' experience in consecutive sessions. Moreover, despite the used by-item and by-subject random intercepts and slopes, we see a high variability of post-editing speed across items and subjects, which makes it difficult to prove significance for the effect of online NMT adaptation for the reduction of post-editing time. This result confirms similar findings reported in \cite{bentivogli2016evaluation}. However, as is shown in the next section, our correlation analysis allows to establish a strong tie between post-editing time and metrics of post-editing effort. 

In order to measure technical effort of post-editing, we combined keyboard strokes and mouse clicks into the metric of key-stroke and mouse-action ratio (KSMR) - a measure proposed by \cite{barrachina2009statistical}. KSMR is calculated as the sum of the number of keystrokes and the number of mouse movements plus one, divided by the count of characters in the reference. We observed significant reduction in KSMR. According to the LMEM analysis, online NMT adaptation enables post-editors to use 12 keystrokes and mouse clicks less per sentence. 

Table \ref{tab:patent} shows example patents which were post-edited during our experiment. In the first example, our adapted NMT system has learned the right translation of \emph{blades} after a post-editor changed \emph{Schaufeln} to \emph{Klingen}. In the second example, the technical term  \emph{image recorder} was inconsistently and wrong translated by the baseline system as \emph{Bildaufzeichner} or \emph{Bildaufnahmeer}. The adapted system learns the translation \emph{Bildaufnahmeapparat} from the post-edit.

\begin{table}[p]
\caption{Examples for test patent data: source, reference, NMT output, post-edit, and adapted NMT output}
\label{tab:patent} 
\begin{center}     
\begin{tabular}{ll}
\hline\noalign{\smallskip}
\multirow{5}{*}{Source} & The outer surfaces of the \textbf{blades} (172) are inclined relative \\
& to the axis of rotation. \\
& When the product to be cut is pushed into the knife arrangement \\
& (170), the latter is rotated in such a manner that the \textbf{blades} (172) \\
& cut the product to be cut along helical paths. \\ 
\noalign{\smallskip}\hline\noalign{\smallskip}
\multirow{5}{*}{Reference} & Die Aussenflächen der Klingen (172) sind relativ zur Drehachse \\
& geneigt. \\
& Wenn das Schneidgut in die Messeranordnung (170) eingeschoben \\
& wird, wird diese derart in Drehung versetzt, dass die Klingen (172) \\
& das Schneidgut entlang helikaler Bahnen zerteilen. \\
\noalign{\smallskip}\hline\noalign{\smallskip}
\multirow{4}{*}{NMT output} & Die Aussenflächen der \emph{Schaufeln} (172) sind zur \\
& Drehrichtungsachse geneigt. \\
& Beim Eindringen des zu durchtrennenden Produkts in die \\
& Messeranordnung (170) wird das zu schneidende Gut so gedreht. \\
\noalign{\smallskip}\hline\noalign{\smallskip}
\multirow{5}{*}{Post-edit} & Die Außenflächen der \textbf{Klingen} (172) sind bezogen auf die \\
& Rotationsachse neigbar. \\
& Beim Einführen des zu schneidenden Produktes in die \\
& Messergruppe (170) wird letztere so gedreht, dass die Klingen \\
& (172) das zu schneidende Produkt spiralförmig schneiden. \\
\noalign{\smallskip}\hline\noalign{\smallskip}
\multirow{5}{*}{Adapted NMT} & Die Aussenflächen der \emph{Schaufeln} (172) sind bezogen auf die \\
& Rotationsachse neigbar. \\
& Beim Einschnitt des Warengutes in die Messeranordnung \\
& (170) wird die \textbf{Klinge} so gedreht, dass die \textbf{Klinge} (172)\\
& das zu schneidende Gut schneidet. \\
\noalign{\smallskip}\hline
\hline\noalign{\smallskip}
\multirow{7}{*}{Source} & Miniaturized \textbf{image recorder} \\
& The aim of the invention is to create an \textbf{image recorder} which \\
& can be miniaturized to such an extent that an endoscope of\\
& the smallest dimensions can be produced therewith. \\
& The disclosed \textbf{image recorder} can also be optimized in such a \\
& manner that the major part of the semiconductor surface is \\
& made available for pixel integration. \\
\noalign{\smallskip}\hline\noalign{\smallskip}
\multirow{7}{*}{Reference} & Miniaturisierter Bildaufnehmer \\
& Ziel der offen gelegten Erfindung ist es einen Bildaufnehmer\\
& vorzuschlagen, der soweit miniaturisiert werden kann, dass \\
& damit ein Endoskop kleinster Dimensionen realisiert werden kann. \\
& Ausserdem kann der offen gelegte Bildaufnehmer dahingehend\\
& optimiert werden, dass der grösste Teil der Halbleiterfläche \\
& für die Integration von Bildpunkten zur Verfügung steht. \\
\noalign{\smallskip}\hline\noalign{\smallskip}
\multirow{7}{*}{NMT output} & Miniaturisierter \emph{Bildaufzeichner} \\
& Der Erfindung liegt die Aufgabe zugrunde, einen \\ 
& \emph{Bildaufnahmeer} zu schaffen, der so miniaturiert ist, dass \\
& eine Impedanz der kleinsten Abmessungen erzielbar ist. \\
& Der \emph{Bildaufzeichner} kann auch so optimiert werden, dass der\\
& grösste Teil der Halbleiterfläche für die galvanische Integration\\
& bereitgestellt wird. \\
\noalign{\smallskip}\hline\noalign{\smallskip}
\multirow{7}{*}{Post-edit} & Miniaturisierter \textbf{Bildaufnahmeapparat} \\
& Das Ziel der Erfindung ist es, einen Bildaufnahmeapparat \\
& herzustellen, der so miniaturisiert ist, dass ein Endoskop \\
& der kleinsten Abmessungen damit hergestellt werden kann. \\
& Der verlautbarte Bildaufnahmeapparat kann auch so optimiert\\
& werden, dass der grösste Teil der Halbleiterfläche für die\\
& Pixelintegration bereitgestellt wird. \\
\noalign{\smallskip}\hline\noalign{\smallskip}
\multirow{7}{*}{Adapted NMT} & Miniaturisierter \emph{Bildlautsprecher} \\
& Der Erfindung liegt die Aufgabe zugrunde, einen\\
& \textbf{Bildaufnahmeapparat} zu schaffen, der so miniaturiert ist, dass\\
& eine Impedanz der kleinsten Abmessungen hergestellt werden kann. \\
& Der \textbf{Bildaufnahmeapparat} kann auch so optimiert werden, \\
& dass der grösste Teil der Halbleiterfläche für die galvanische\\
& Integration bereitgestellt wird. \\
\noalign{\smallskip}\hline
\end{tabular}
\end{center}
\end{table}

\subsection{Correlation Study}

\begin{figure}[t]
	\begin{minipage}{0.48\textwidth}
     	\centering
  		\includegraphics[width=0.85\textwidth, page=1]{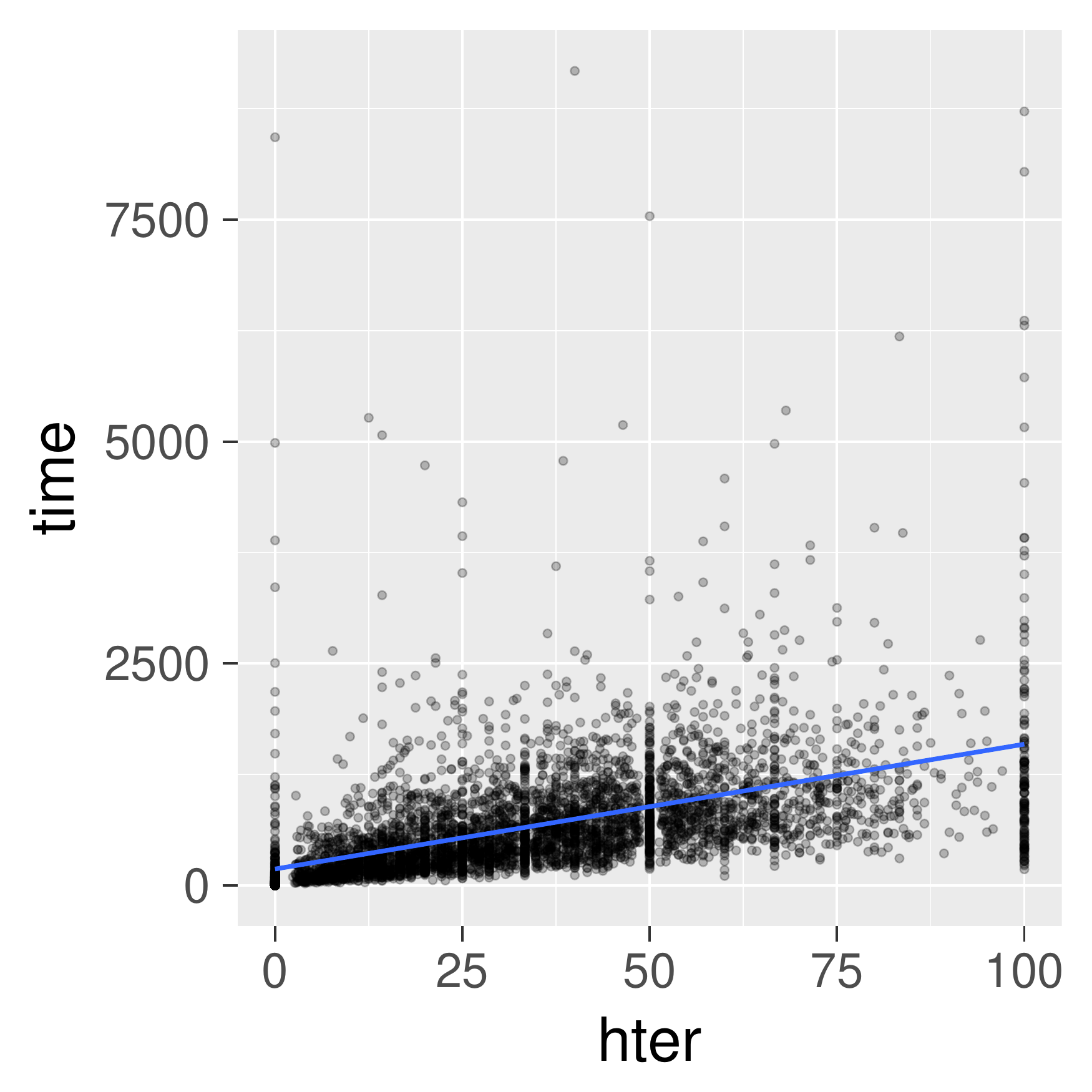}
  	\end{minipage}\hfill
   	\begin {minipage}{0.48\textwidth}
     	\centering
  		\includegraphics[width=0.85\textwidth, page=2]{cor2_time_hter_hbleu.pdf}
  	\end{minipage}
	\caption{Regression plots of hTER and time (left) and hBLEU and time (right)}
	\label{fig:hter_hbleu_time}      
\end{figure}

Since due to the relatively small size of our user study, significance of the result difference between the adaptation conditions could not be established for some response variables, we furthermore analyzed the correlations between the obtained measures. For this purpose, we calculated pairwise Pearson correlation coefficients (Figure \ref{fig:corr_matrix_P}). The correlation matrix shows strong correlations of translation quality rating with hBLEU, hTER, time and KSMR. Figure \ref{fig:hter_hbleu_time} shows detailed values for post-editing time plotted against hTER and hBLEU. The plots confirm that while the effect of online adaptation was not significant for the response variable of time, there is a strong correlation of post-editing time with metrics for post-editing effort such as hTER or hBLEU.

\section{Summary and Conclusion}
\label{conclusion}
We presented a user-study on the effect of online adaptation on NMT systems to interactive user post-edits of the proposed translations. We found significant reductions in human post-editing effort along several well-established response variables (hTER, hBLEU, KSMR).
Furthermore, we found a domain adaptation effect due to online adaptation, leading to significant improvements of TER/BLEU of the machine translations with respect to human reference translations. This provides further evidence for the advantages of an online adaptive NMT system in an interactive setup.

Due to our experimental setup where the same documents were translated by both a static and an adaptive NMT system, and post-edited by two different translators at different points in time, we found a confounding effect between improved post-editing experience and reduced time. This did not allow us to establish significant improvements of online NMT adaptation with respect to post-editing time. However, we found a strong correlation of reduction in post-editing time to improvements in metrics for post-editing effort such as hTER, hBLEU, or KSMR. This shows firstly that the latter metrics are more reliable indicators for reduced post-editing effort, and furthermore that reduced post-editing time is a correlated, but not necessarily directly causally related effect.

In sum, our user study established significant improvements due to online NMT adaptation along well-known metrics of post-editing effort, and along the dimension of domain adaptation that is particularly important in technical translation domains such as patent translation. Our user study did not touch novel modes of user interaction with NMT systems \citep[for example, human bandit feedback:][]{KreutzerETAL:17,NguyenETAL:17} or alternative modes of NMT system adaptation \citep[for example, interactive translation prediction:][]{KnowlesKoehn:16}. These topics are subject of ongoing and future work \citep[see][]{KreutzerETALnaacl:18,KreutzerETALacl:18,LamETAL:18}.

\begin{acknowledgements}
  The research reported in this paper was supported in part by the German research foundation (DFG) under grant RI-2221/4-1. 
\end{acknowledgements}

\bibliographystyle{spbasic}       
\bibliography{references}   

%
%

\end{document}